\documentclass[letterpaper, 10pt, conference]{ieeeconf}
\IEEEoverridecommandlockouts
\usepackage{cite}
\usepackage{amsmath,amssymb,amsfonts}
\usepackage{algorithm}
\usepackage{algorithmic}
\usepackage{graphicx}
\usepackage{textcomp}
\usepackage{xcolor}
\usepackage{booktabs}
\usepackage{array}
\usepackage{multirow}
\usepackage{url}
\usepackage{tikz}
\usetikzlibrary{arrows.meta,positioning,shapes.geometric}
\usepackage[caption=false,font=footnotesize]{subfig}
\makeatletter\let\labelindent\@undefined\makeatother
\usepackage{enumitem}
\usepackage{float}
\usepackage[colorlinks=true, citecolor=blue, linkcolor=blue, urlcolor=blue]{hyperref}

\def\BibTeX{{\rm B\kern-.05em{\sc i\kern-.025em b}\kern-.08em T\kern-.1667em\lower.7ex\hbox{E}\kern-.125emX}}

\newcommand{\Lzero}{$L_0$}
\newcommand{\Lone}{$L_1$}
\newcommand{\Ltwo}{$L_2$}
\newcommand{\eg}{e.g.,\ }

\begin{document}

\title{\LARGE\bf Diagnosing and Recovering from Observation-Space Shift at Long-Horizon Skill Seams}

% ---- Double-blind toggle: \reviewtrue for anonymous submission (ICRA is
% double-blind) \reviewfalse for the camera-ready with real names. -----------
\newif\ifreview \reviewfalse
\ifreview
  \author{Anonymous Author(s)\thanks{Paper \#XXXX. Submitted to ICRA 2027 under
  double-blind review; code and artifacts available upon acceptance.}}
\else
  \author{Pranav Wagh$^{1}$, Yu Fang$^{1}$, Yue Yang$^{1}$, and Mingyu Ding$^{1}$%
  \thanks{$^{1}$Department of Computer Science, University of North Carolina at
  Chapel Hill, NC, USA. {\tt\small\{pawagh, yufang, yygx, md\}@cs.unc.edu}}}
\fi
\maketitle
\thispagestyle{empty}\pagestyle{empty}

% ============================================================================
\begin{abstract}
% problem -> diagnosis -> why existing methods fail (challenges) -> our method (modules) -> experiments, results
% ============================================================================

Long-horizon robotic manipulation is often built by chaining skills that are trained independently.
Although individual skills can be reliable in isolation, long-horizon task performance can degrade sharply when these skills are chained: each downstream skill must start from the state produced by its predecessor rather than from its training distribution.
We study this failure mode, known as Observation-Space Shift (OSS), and investigate what aspects are responsible for such skill-seam failures.
Through privileged simulator resets, we identify that the dominant shift arises from displaced scene state (e.g., an open drawer or secondary objects left behind by preceding skills), rather than from the robot joint configuration or the object manipulated by the downstream skill.
To test this diagnosis, we build a fully learned \emph{detect–restore–resume} system that restores the components the analysis identifies: a task-progress monitor \emph{detects} the stall, a learned policy \emph{restores} those components, and seam-robust fine-tuning lets the skill \emph{resume}. 
It recovers the seam where every tested alternative fails, which is evidence for the diagnosis rather than a general purpose method.
On the BOSS-44 benchmark, our system improves full-chain success from 7.6\% to 26.5\%, a $3.5\times$ improvement over the base policy and 51\% of a privileged restoration oracle.
In contrast, best-of-$K$ resampling, a Diffusion Policy, and world-model-based baselines fail to recover from the evaluated OSS seam states.
During real-robot experiments on a real Franka arm running $\pi_{0.5}$ policy, the same progress monitor is limited by external camera observability, yet closing the loop still recovers some otherwise-terminal failures. This motivates wrist and gripper sensing.
These results suggest that some long-horizon composition failures are better addressed by restoring the scene before resuming the policy, rather than retrying from an off-support state.
\end{abstract}

% ============================================================================

\section{Introduction}
% ============================================================================

Long-horizon robotic manipulation often requires composing multiple skills into a single task~\cite{liu2023libero,yang2025boss}.
For example, a robot may first open a drawer, then move an object, and finally place it at a target location.
A common strategy is to train individual manipulation skills independently and execute them sequentially in a chain at deployment.
Although these skills can be reliable when evaluated from their own training distributions, long-horizon task performance can degrade drastically once they are chained together.

The challenge arises at the \emph{skill seams}, where one skill terminates and the next begins:
A downstream skill does not start from the canonical initial states on which it was trained;
instead, it inherits whatever scene state its predecessor leaves behind.
For example, an ``open drawer'' skill may leave the drawer open before a ``place bowl" skill, trained with the drawer closed, begins execution.
The downstream skill is therefore asked to act from an unfamiliar state and may stall even though it succeeds reliably in isolation.
This mismatch between the downstream skill's training distribution and the state encountered during skill composition is known as \textbf{Observation-Space Shift (OSS)}~\cite{yang2025boss}.
On the three-skill chains studied in BOSS-44~\cite{yang2025boss}, success through the chain drops from $88.5\%$ after executing the first skill, to $34.3\%$ after two skills (one seam), and to $7.6\%$ after all three skills (two seams), despite the component skills achieving $79.0\%$ average success when evaluated independently from their canonical initial states.

% ---- EYECATCHER: single-column, top of right column 1 next to the abstract 
% ============================================================================
\begin{figure}[t]
\centering
\includegraphics[width=\columnwidth]{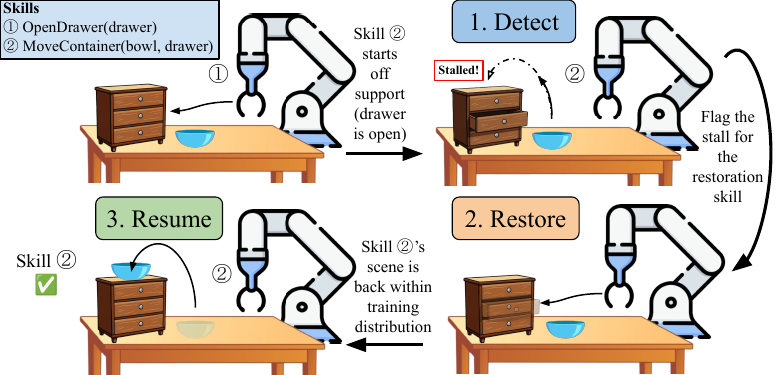}
\caption{\textbf{Detect, restore, resume at a skill seam.} A chained skill collapses when it starts from the previous skill's off-support terminal state; a task-progress monitor detects the stall, a learned restoration skill returns the displaced scene to support, and the chain resumes.}
\label{fig:teaser}
\end{figure}

% ============================================================================

What changes at a skill seam that causes the downstream skill to fail?
We consider three potential sources of shift: the preceding skill may
(1) leave the robot arm in an unfamiliar state;
(2) displace the object manipulated by the downstream skill;
or (3) alter other elements of the scene.
To identify which components are responsible, we diagnose OSS through privileged simulator interventions.
Using a teleportation oracle and component-wise resets, we selectively restore different state components to their canonical values.
We find that
(1) resetting the robot arm provides little improvement;
(2) the object manipulated by the downstream skill is typically already close to its canonical position;
and (3) the dominant recovery comes from restoring articulated scene state (e.g., an open drawer) and secondary objects (i.e., other objects displaced by preceding skills but not manipulated by the downstream skill).
Moreover, these effects are complementary: restoring articulation or secondary objects individually yields only partial gains, whereas restoring them jointly recovers substantially more of the oracle performance.

This diagnosis suggests that recovery should modify the state responsible for the failure rather than repeatedly invoke the policy from off-support states.
As shown in Fig.~\ref{fig:teaser}, we introduce a \textbf{detect–restore–resume} framework: a fully learned, non-privileged system that recovers much of the oracle from cameras and proprioception alone.
A \emph{task-progress monitor} detects when execution has stalled.
A \emph{learned restoration policy} modifies the displaced scene state, after which the downstream skill resumes execution.
Since learned restoration only approximately returns the scene to its canonical state, we additionally apply \emph{seam-robust prevention fine-tuning} on post-restoration hand-off states to improve robustness during re-entry.
On BOSS-44, our system improves three-skill full-chain success from $7.6\%$ to $26.5\%$, compared with $51.8\%$ under a privileged oracle that exactly restores the scene to its canonical state.
A matched re-home-and-retry baseline reaches only $12.9\%$, showing that the improvement cannot be explained simply by additional execution attempts.
Best-of-$K$ resampling and a Diffusion Policy trained on the same canonical demonstrations also fail from the evaluated OSS seam states, indicating that retrying or changing the policy architecture alone does not resolve these failures.
On a real Franka arm with a $\pi_{0.5}$ policy, the same monitor runs unchanged but transfers only weakly from an exterior camera. Closing the loop also recovers some otherwise-terminal failures, a proof of concept that isolates the observability (wrist and gripper sensing) as an important requirement for deployment.

Our contributions are three-fold.
\begin{itemize}[leftmargin=*, noitemsep, topsep=0pt]

\item We provide a diagnosis of OSS at long-horizon skill seams. Through privileged simulator interventions, we show that skill-seam failures are driven primarily by displaced articulation and secondary scene objects.

\item We propose a learned detect–restore–resume framework for long-horizon skill composition. The system detects stalled execution, restores the displaced scene toward the downstream support, and resumes execution using only deployable observations at test time.

\item Experiments on both BOSS-44 demonstrate that our framework achieves a $3.5\times$ improvement in full-chain success, outperforming other baselines. We provide empirical analysis of when restoration is a better intervention than retry. On a real Franka, the same monitor runs unchanged but it limited by exterior-camera observability. Closing the loop recovers a fraction of otherwise-terminal failures, a proof of concept that motivates wrist and gripper sensing.  

\end{itemize}

% ============================================================================
\section{Related Work}
% ============================================================================

\noindent \textbf{Skill chaining, observation-space shift, and recovery.}
Composing pretrained skills into long-horizon behavior is central to hierarchical and library-based control~\cite{sutton1999options,garrett2021tamp,yang2025boss,liu2023libero}, where a transition succeeds only when one skill's termination region meets the next's initiation set~\cite{konidaris2009chaining,bagaria2020dsc}. A seam is a violation of that condition. The mechanism is the behavior-cloning covariate shift at composition, where a downstream skill inherits a state its predecessor produced but never trained on~\cite{ross2011dagger,laskey2017dart}. This is not specific to small policies: generalist vision-language-action models~\cite{brohan2023rt1,zitkovich2023rt2,kim2024openvla,octo2024,black2024pi0} and modern imitation learners~\cite{florence2021ibc,mandlekar2021robomimic,diffusionpolicy} clone actions and inherit the same off-support behavior when chained. Unlike work that reports chaining loss as an aggregate drop, we localize the shift causally to specific state components. Prior fixes act at training time, regularizing a skill's terminal distribution toward the next initiation set~\cite{leelim2021terminal} or appending recovery and correction data~\cite{rac}. Test-time recovery instead retrieves and replays demonstration sub-trajectories~\cite{strap}, backtracks to a prior skill, takes object-centric keypoint steps back onto the manifold~\cite{ocr2024}, or learns local RL recovery policies~\cite{recoverychaining}. We test faithful instances of retrieval, backtracking, and library-inverse correction and find none re-manipulates the displaced scene. Our restoration skill supplies that behavior as a learned reset controller, the objective of reset-free RL and learned resets~\cite{eysenbach2018leavenotrace,gupta2021resetfree,sharma2022earl}, but invoked at a detected seam rather than every episode. It is trained with SAC~\cite{sac} and RLPD~\cite{rlpd} from real failure states and distilled to a camera-only student. We compare it against a training-time prevention condition built with data generation~\cite{mimicgen} rather than only asserting the distinction.

\noindent \textbf{Detection and test-time alternatives.}
Detecting task failure without labels has converged on task-progress and value estimation from pixels: Generative Value Learning~\cite{ma2025gvl} elicits per-frame completion from a VLM, and process-reward models such as Robometer~\cite{robometer} train per-frame progress on robot data. Our monitor keeps the progress-as-normalized-time target but trains a $\sim\!1$M-parameter head over a frozen encoder with an added success/stall contrast loss, and fires a calibrated stagnation trigger in the spirit of distribution-free failure detection~\cite{safe,farid2022ood}. Unlike runtime monitors that add temporal-consistency terms~\cite{agia2024sentinel} or world-model prediction surprise, we find progress alone is the correctly signed signal at stalls. The alternative to restoration is planning or search: imagine-score-execute pipelines plan with learned dynamics~\cite{ha2018worldmodels,hafner2023dreamerv3,tdmpc2,kim2026cosmos} and sampling-based shooting (\eg MPPI~\cite{mppi}) selects actions by simulated return, while a separate line scales test-time compute by sampling $K$ rollouts and keeping the best~\cite{snell2024scaling}. Evaluated from OSS seams, a world model as generator, monitor, or planner gives no benefit or is falsified, and best-of-$K$ cannot select what the policy cannot produce. These both point to restoration over planning or sampling.

% ============================================================================
\section{Problem Formulation}
\label{sec:problem}
% ============================================================================

\textbf{Setting.}
We study three-skill chains on BOSS-44, a benchmark of $44$ kitchen and living-room manipulation tasks drawn from LIBERO-90~\cite{liu2023libero}.
Tasks are simulated in robosuite~\cite{robosuite2020} on the MuJoCo physics engine~\cite{mujoco2012}, with a $7$-DOF Panda under $20$~Hz OSC delta control.
A chain $\mathrm{ch3}\_N$ is an ordered triple of BOSS-44 tasks run in one continuous
episode. For example, $\mathrm{ch3}\_2$ is \emph{open the drawer $\to$ place the bowl
on the plate $\to$ place the bowl in the cabinet}. All ten catastrophic chains are
reorderings of the shared $13$-skill set. When skill~$i$'s goal is satisfied, the environment advances to skill~$i{+}1$, re-homing only the arm (joint and velocity state) while objects and articulation carry over.
That carry-over is what we study.
A three-skill chain is the shortest that contains two seams, so a downstream skill can inherit shift that has already compounded across an upstream one. It is the minimal setting in which seam collapse is more than a single-transition artifact, and we adopt it throughout.
Each skill is a per-task BC-Transformer policy (one checkpoint per task, frozen throughout) reading a fixed exterior camera (``agentview''), wrist RGB, and $16$-D proprioception (arm joint and gripper state). We report chain success at three depths: \Lzero{} (first skill only), \Lone{} (through the first seam, two skills), and \Ltwo{} (all three skills, the full chain and our primary metric).

\textbf{Observation-Space Shift.}
Observation-Space Shift (OSS) is the covariate shift a chained skill faces at a seam~\cite{yang2025boss}.
% We restate it here in the notation we use.
Let $\mathcal{D}_i$ be the canonical initial-state distribution skill~$i$ was trained from.
In a chain, skill~$i$ is executed from $s \sim T_{i-1}$, the terminal-state
distribution of its predecessor.
\emph{Seam collapse} is the failure that occurs because $T_{i-1}$ is a covariate shift off $\mathcal{D}_i$'s support, and BC provides no off-support correction.
We make this quantitative below.

\textbf{Scope of restoration.} In TAMP terms, the effects of skill~$i{-}1$ split into
\emph{preconditions} of skill~$i$ and \emph{side effects} unrelated to it. Recovery by
restoration is well defined only for side effects, and only when the change is
reversible toward a known canonical target. The inherited change is itself a
precondition, like opening a drawer before placing an object \emph{in} it, and restoring it
destroys what the next skill needs. Sec.~\ref{sec:whenrestore} maps this boundary
empirically.

\textbf{Benchmark subset and evaluation.}
We evaluate $20$ episodes per chain, and extend each estimate to five disjoint
seeds by sliding the init-state window.
% (offsets $\{0,20,40,60,80\}$; identical protocol, disjoint initial conditions).
Our primary analyses focus on the $10$ catastrophic chains (base-policy \Ltwo{}$\,\le 0.30$), which exercise $13$ of the $44$ skills across three LIBERO-90 kitchen scenes whose seams leaves a drawer open S1 (top/bottom drawers, a black bowl, a plate), S2 (top drawer, black bowls, a plate), S4 (bottom drawer, a wine bottle and bowl, where the object grasp is the hardest). Seam collapse is most severe.
The collapse is noticeably seen on some chains where chain success falls $0.885 \to 0.343 \to 0.076$ across the two seams, with each skill individually competent: evaluated from their own canonical initial states the frozen skills succeed at $0.79$ on average (Table~\ref{tab:isolation}), so the collapse is composition, not skill incompetence.
Across the full benchmark, OSS induces a mean Ratio Performance Delta (RPD, the relative success drop) of $0.45$ over one seam and $0.51$ over two~\cite{yang2025boss}; the catastrophic chains are the severe tail of that distribution.

\begin{table}[t]
\centering\footnotesize\setlength{\tabcolsep}{4pt}
\caption{Per-skill success in isolation ($L_0$), catastrophic-chain skills. Mean $0.79$ (range $0.35$--$1.00$): the skills are competent alone, so chain collapse is a composition failure.}
\label{tab:isolation}
\begin{tabular}{l l c @{\hskip 12pt} l l c}
\toprule
Sc & Skill & $L_0$ & Sc & Skill & $L_0$ \\
\midrule
S1 & open top drawer      & $1.00$ & S2 & mid bowl on plate       & $0.45$ \\
S1 & open bottom drawer   & $0.90$ & S2 & mid bowl on cab.\ top   & $0.90$ \\
S1 & bowl on plate        & $1.00$ & S4 & close bottom drawer     & $1.00$ \\
S1 & bowl on cab.\ top    & $0.70$ & S4 & bowl in bottom drawer   & $0.80$ \\
S2 & open top drawer      & $1.00$ & S4 & bowl on cab.\ top       & $0.95$ \\
S2 & bowl (back) on plate & $0.60$ & S4 & wine in bottom drawer   & $0.65$ \\
\multicolumn{3}{c}{} & S4 & wine on wine rack & $0.35$ \\
\bottomrule
\end{tabular}
\end{table}

% ============================================================================
\section{Method: Detect, Restore, Resume}
\label{sec:method}
% ============================================================================
Recovering a chained skill requires noticing it was stalled, putting the scene back on-support, and letting the skill tolerate an imperfect hand-back. We add one learned component for each, \emph{detect}, \emph{restore}, and \emph{resume}, leaving the base skills frozen. All three read only deployable inputs (RGB cameras and proprioception; Fig.~\ref{fig:teaser}).

\subsection{Detect: a task-progress monitor with a calibrated trigger}
\label{sec:detect}
\begin{figure}[t]
\centering
\includegraphics[width=0.33\textwidth]{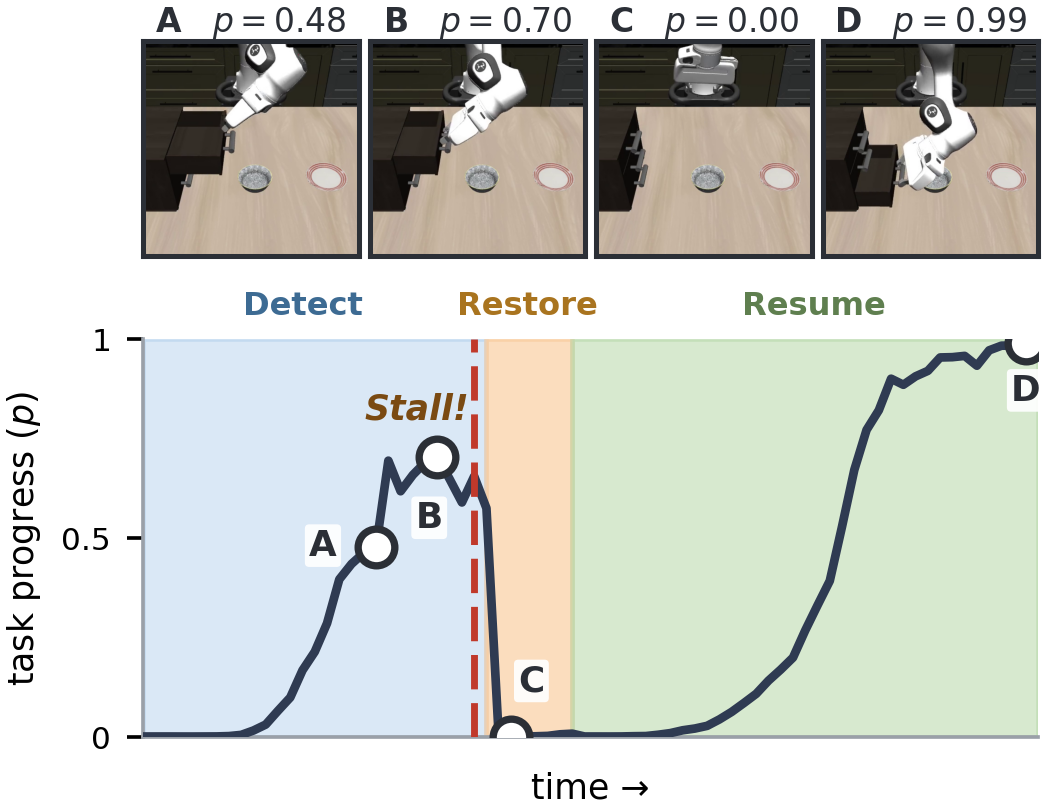}
\vspace{-12pt}
\caption{\textbf{The task-progress monitor in action (simulation, S1).} Frames A--D from one execution with their monitor scores: reaching ($p{=}0.48$), stalled at the seam ($p{=}0.70$), reset to the seam by restoration ($p{=}0.00$), and completed ($p{=}0.99$). Below, the same scores over time: progress climbs, stalls (\emph{Stall!}), the calibrated trigger fires (dashed), restoration drops it to zero, and the retry resumes. Monitor trigger $+$ learned restoration, no oracle. Phases
colored as in Fig.~\ref{fig:teaser}.}
\label{fig:trace}
\end{figure}

Detect answers one question: has the current skill stopped making progress?
A task-progress head scores each frame's position within a skill, and the trigger fires when that score stops rising (Fig.~\ref{fig:trace}).
Prediction-surprise signals fail here: at seams the robot stalls, so the next
frame is nearly identical and trivially predictable, and surprise becomes
anti-predictive ((Sec.~\ref{sec:controls})). A stalled robot cannot manufacture
progress, so the progress signal stays correctly signed.

\textbf{Trigger.} The alarm is progress-slope stagnation over an $8$-sample window
(not absolute level, which is low early in every skill), fired when the statistic
exceeds a threshold $\tau$ for $3$ consecutive samples. 
$\tau$ is set by split-conformal calibration on success demos~\cite{safe}. Since seam states are off-support its formal guarantee need not hold, so we fix the operating point empirically on cached rollouts ($0.95$ true-positive, $7\%$ false-alarm;
Sec.~\ref{sec:deploy}).

\subsection{Restore: a learned scene-restoration skill}
\label{sec:restore}
Restoration must return the components the diagnosis singles out, like the displaced articulation and the secondary objects prior skills leave behind (Sec.~\ref{sec:diagnosis}), to where the next skill is competent. No existing action source does this, and four candidates (a scripted controller, demonstration replay, backtracking, and a library inverse skill) all fail since skill libraries encode task progress, not scene restoration (Sec.~\ref{sec:controls}). We therefore \emph{learn} the behavior with RL under a dense reward.

\textbf{Integration.} At an alarm the restoration actor runs closed-loop until the
drawer returns to canonical ($\sum|\Delta_{\text{artic}}|<0.02$) or a $150$-step cap
fires. The arm then re-homes and the skill resumes with a cleared observation history.
Up to $r_{\max}{=}2$ alarms are handled per skill.

\textbf{Deployability.} The state-based teacher is distilled to a camera-only student
(dual RGB and $16$-D proprioception; $0.85/0.90$ vs.\ teacher, Fig.~\ref{fig:student}),
so the loop uses zero simulator state. The same controller runs unchanged in simulation
and on hardware behind a minimal robot API (reset, step, observe, home).

\subsection{Resume: seam-robust prevention fine-tuning}
\label{sec:prevent}
Resume makes the downstream skill tolerate an imperfect hand-back, since restoration returns the scene only approximately to support. We fine-tune each post-seam skill on the deep hand-off states the recovery loop reaches, which raw rollouts almost never produce. On its own this is marginal, adding to recovery only at a trend level (Sec.~\ref{sec:main}); we treat it as complementary to restoration.

\subsection{Implementation Details}
\textbf{Detector Architecture.} A frozen DINOv3 ViT-B/16~\cite{dinov3} encodes a
$128{\times}128$ frame to a patch-token grid. A small task-conditioned head (a
FiLM-conditioned~\cite{film} ResNet, $\sim\!1$M parameters) regresses the progress
logit. The head is trained with three self-supervised losses on strong-policy
rollouts: an absolute-progress anchor (BCE to $t/(T{-}1)$, GVL-style~\cite{ma2025gvl}),
an intra-demo ranking loss (later frames score higher), and a success/stall
contrast that pushes frames from failed-window tails below success frames, the
only failure-specific signal available without labels. All numbers are
cross-validated by chain (each chain scored by the fold that never trained on it).

\textbf{Restoration Reset library.} Every monitor alarm records the full simulator state, and
each recorded failure state is paired with its exact restoration target (the
canonical articulation and object values the oracle would teleport to). This
yields per-scene libraries of realistic failure states
($\sim\!5.7$k$/4.6$k$/3.6$k for S1/S2/S4, $\sim\!14$k total) with dense, automatic
targets: no reward engineering, no labels, and every rollout starts in a state
that occurs mid-chain.

\textbf{Restoration Agent.} A render-free, state-based teacher is trained per scene by
SAC~\cite{sac} (drawer S1/S2) and demonstration-augmented RLPD~\cite{rlpd} (drawer
S4, where SAC's grasp exploration stalls). The reward is the distance of the
displaced component to its known target, with a disturbance penalty on non-target
objects and a success bonus. Episodes are short ($\le\!150$ steps) and
success-terminated. State-based training needs no rendering, so it parallelizes on
CPU. Drawer restoration reaches $1.00/0.93/0.78$ (S1/S2/S4) at the skill level
(Fig.~\ref{fig:rlpd}).

\textbf{Prevention fine-tuning.} We fine-tune each post-seam skill on $581$ deep hand-off demonstrations collected through the recovery loop, mixed with the original per-skill demonstrations. We keep the checkpoint that minimizes seam-window loss subject to a forgetting constraint on the base-policy validation loss. Scaling this set $10\times$ (to $6073$) does not improve it (Sec.~\ref{sec:future}).

% ============================================================================
\section{Experiments}
\label{sec:exp}

\begin{figure*}[t]
\centering
\includegraphics[width=\textwidth]{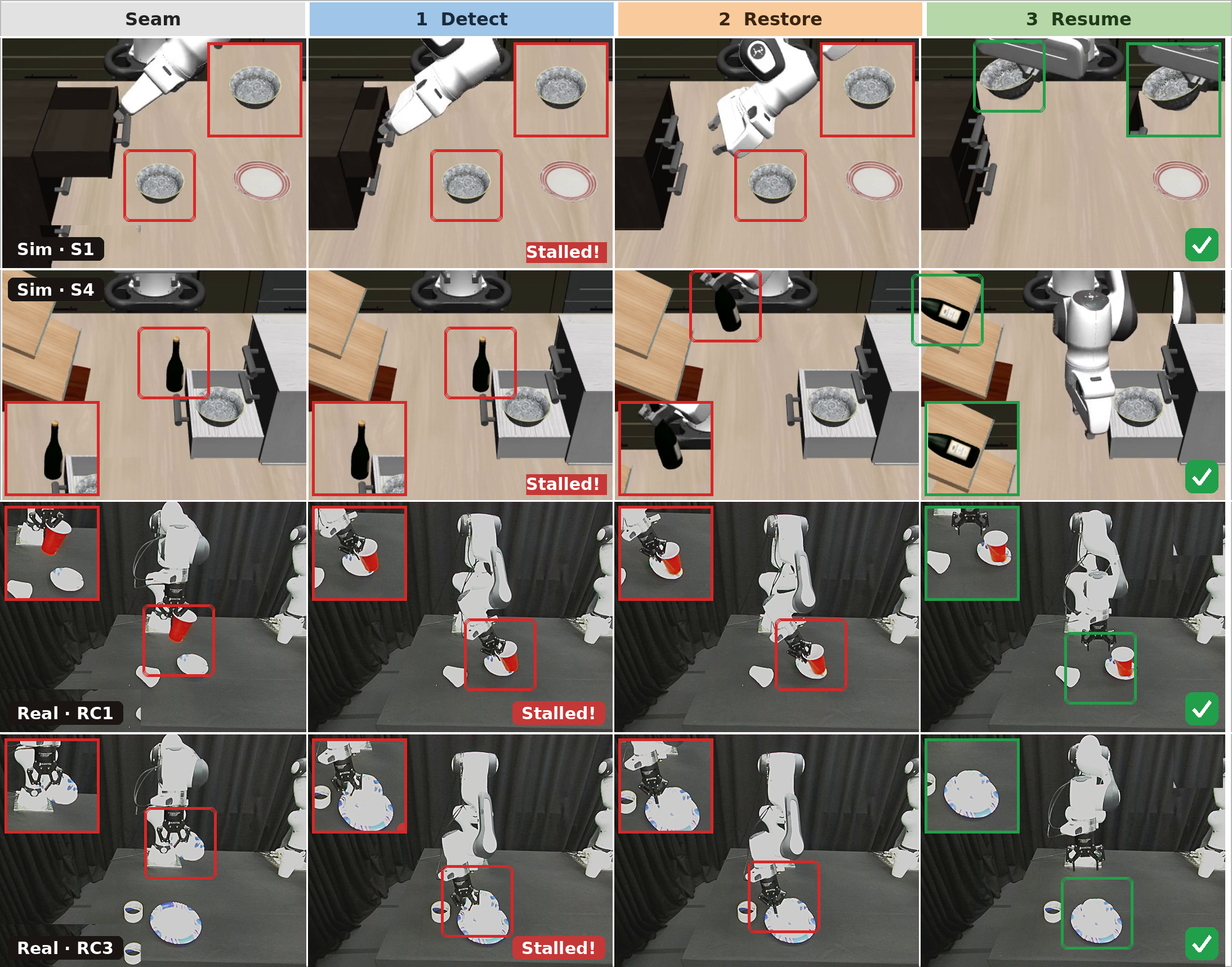}
\caption{\textbf{The detect--restore--resume loop across scenes and on hardware.}
Each row runs one chained skill through a seam, left to right. \emph{Seam}: the
downstream skill begins from the previous skill's off-support terminal state: an open
drawer in simulation, a mid-place hand-off on hardware. \emph{Detect}: the
task-progress monitor fires (``Stalled!''). \emph{Restore}: the learned restoration
skill returns the displaced articulation and objects to the skill's support.
\emph{Resume}: the skill resumes and completes ($\checkmark$). Rows: two catastrophic
simulation scenes (S1, open-drawer then place the bowl; S4, a distinct cabinet scene),
then hardware chains RC1 (cup$\,\to\,$plate) and RC3 (bowl$\,\to\,$plate). 
Without the loop, execution halts at the \emph{Seam} column.
Simulation frames are the full system on the paper's checkpoint
(Table~\ref{tab:main}); hardware shows recovered place attempts
(Sec.~\ref{sec:realrobot}). Each panel boxes the target object in red and shows a magnified inset of it. The box and inset turn green once the object is placed and the skill completes.}
\label{fig:system}
\end{figure*}

% ============================================================================

\textbf{Metrics.} Detection is scored by the AUROC of the alarm statistic over per-seam windows.

\textbf{Overview.} We structure the evaluation by first localizing \emph{what} causes seam collapse and must be restored (Sec.~\ref{sec:diagnosis}).
We then show that no simpler alternative, like test-time compute, a stronger base policy, or a world model, recovers it (Sec.~\ref{sec:controls}).
We then present the deployed detect–restore–resume system and its recovery (Sec.~\ref{sec:main}).
Finally, we analyze \emph{when} restoration is the right action, and why (Sec.~\ref{sec:whenrestore}).
We also quantify detector quality and deployability (Sec.~\ref{sec:deploy}) and test transfer to a real robot (Sec.~\ref{sec:realrobot}).

\subsection{What causes seam collapse, and what must be restored}
\label{sec:diagnosis}
We begin by asking \emph{what} a recovery must put back, since a fix cannot be designed
without it. Using a privileged teleport oracle, we restore one scene component at a
time (only the arm, only the drawer, or only the objects to canonical at the alarm) and
measure the gain ($n{=}100$; Tab.~\ref{tab:ablation}). This coarse diagnostic run has a
base ($0.098{\pm}0.031$) that differs from the $5$-seed base ($0.076$,
Tab.~\ref{tab:main}). We use it for the component decomposition, not the headline.

\begin{table}[t]
\centering
\setlength{\tabcolsep}{10pt}
\caption{Component-wise reset ablation.}
\label{tab:ablation}
\setlength{\tabcolsep}{10pt}
\begin{tabular}{lc}
\toprule
Restored component(s) & \Ltwo{} \\
\midrule
Base (none)                       & $0.098 \pm 0.031$ \\
Arm only                          & $0.132 \pm 0.023$ \\
Arm $+$ articulation              & $0.212 \pm 0.023$ \\
Arm $+$ secondary objects         & $0.268 \pm 0.036$ \\
\textbf{Full (oracle)}            & $\mathbf{0.518 \pm 0.018}$ \\
\bottomrule
\end{tabular}
\par\smallskip
\begin{minipage}{\columnwidth}\footnotesize
$n{=}100$, catastrophic chains. Arm-only matches base. Drawer articulation and
secondary objects each carry about half, and jointly they exceed the sum.
\end{minipage}
\end{table}

Three findings follow. (i)~The robot's own configuration is not the bottleneck:
resetting the arm alone ($0.132$) is within noise of base ($0.098$), which rules
out ``bridge the robot state back'' explanations. (ii)~Supporting triage shows the
object the next skill manipulates is essentially canonical (median displacement
$1$~cm, and $25$~cm from the gripper), and nothing is ever physically lost
($2/10{,}036$ states dropped, $0.02\%$), so the deployable ceiling is not
physically capped. (iii)~The gain lives in drawer articulation and secondary
objects, and their effects are complementary: the individual increments sum to a
predicted $0.35$, but joint restoration yields $0.52$, an excess of
$0.170\pm0.052$ (propagated SEs, ${\approx}3.3\sigma$). Partial fixes do not stack;
the scene must be restored jointly. This is precisely the specification the learned
restoration skill (Sec.~\ref{sec:restore}) targets.

\subsection{No tested alternative recovers the seam}
\label{sec:controls}
The diagnosis says the \emph{state} must be restored, but perhaps a cheaper
intervention can address the issue. 
If seam collapse were merely bad luck (an unlucky rollout), weak base skills (a better policy would generalize), or a planning failure (a world model would fix it), a less costly fix than restoration should work. 
We tested each in its strongest form from the actual seam states, and all fail.

\begin{table}[t]
\centering\footnotesize\setlength{\tabcolsep}{4pt}
\caption{No tested alternative recovers the seam.}\label{tab:controls}
\begin{tabular}{ll}
\toprule
Alternative & Result at the OSS seam \\
\midrule
Best-of-$K$ ($K{=}8$)          & $0.00$ \; (clean $0.35\!\to\!1.00$) \\
Diffusion Policy ($2500$ ep)   & $0.00$ \; (clean $\le\!0.70$) \\
World model (generator)        & $\approx$ base \\
World model (monitor)          & AUROC ${<}0.5$ \\
World model (planner)          & $0.65$ vs.\ MB $0.53$ \\
End-to-end RL~\cite{recoverychaining} & $0.45$ vs.\ $0.52$ \\
\midrule
\textbf{Restoration (ours)}    & $\mathbf{0.265}$ \\
\bottomrule
\end{tabular}
\end{table}

\noindent \textbf{Test-time compute cannot help} (our strongest control). From a clean init, at least one of $K{=}8$ rollouts succeeds, so monitor selection is trivially perfect (Tab.~\ref{tab:controls}). From the mid-chain seam the base policy fails on all eight, leaving nothing to select.

\noindent \textbf{A stronger base policy does not help.} A faithful Diffusion
Policy~\cite{diffusionpolicy} (canonical conditional U-Net, $2500$ epochs) succeeds
from canonical initial states but matches BC exactly from seam states
(Tab.~\ref{tab:controls}), and is non-uniform (one task worsens with more epochs).
This tests a stronger policy \emph{architecture} on the same canonical demonstrations,
not a base policy trained on seam data (the prevention route in Sec.~\ref{sec:main}).

\noindent \textbf{A world model does not help in any of its three roles} (pre-registered
ablation; Tab.~\ref{tab:controls}). As a generator, splicing predicted frames across
the seam ties baseline (the policy's CNN treats generated pixels as out-of-distribution).
As a monitor, prediction surprise inverts on stalls, which motivates our progress
monitor. As a planner, model-free RLPD exceeds model-based DEMO3/TD-MPC2~\cite{tdmpc2}
at $3\times$ the sample efficiency. Even a perfect world model, simulator-scored
shooting over arm primitives, fails. This confirms that the binding constraint is the
action repertoire, which model quality does not supply.

\noindent \textbf{End-to-end RL recovery underperforms modularity.} A single policy trained
directly on whether the next skill succeeds (RecoveryChaining~\cite{recoverychaining}
with RLPD) stays below the demonstration baseline (Tab.~\ref{tab:controls}). The
decomposed design of a specialist restorer plus robust skills outperforms it.

Together these isolate the cause. More sampling (best-of-$K$) and a stronger policy \emph{architecture} (Diffusion Policy) both fail from the seam, so the gap is not test-time search or policy capacity. 
Scaling seam demonstrations $10\times$ within the same source distribution leaves prevention unchanged (\Ltwo{}~$0.117$; Sec.~\ref{sec:limitations}), while restoring the state recovers it. Within the distribution we sampled, added data does not substitute for restoration
This does not test every conceivable fix (see
Sec.~\ref{sec:limitations} on training-time data scaling).

\subsection{The detect–restore–resume system recovers the collapse}
\label{sec:main}

Having localized the cause (Sec.~\ref{sec:diagnosis}) and ruled out simpler fixes (Sec.~\ref{sec:controls}), we test the diagnosis directly. If correct, it recovers most of the collapse, and Table~\ref{tab:main} makes three points showing why it works.
First, \emph{restoration} drives the gain: against a base given the same re-home/retry budget (\Ltwo{}\,$0.129$) the full system still adds $+0.136$ ($2.05\times$), so the effect is not just added steps. 
This is greater than the gain that extra compute or prevention provide. 
Second, it reaches $51\%$ of the privileged oracle, so roughly half the recoverable gap remains,
and that remainder is restoration \emph{quality}, not detection. 
Third, neither component is significant alone (recovery $p{=}0.094$, prevention $p{=}0.188$), so the
claim is at the system level. 
Table~\ref{tab:main} reports the four-condition ablation plus the oracle on the $10$ catastrophic chains (statistics in caption).

\begin{table}[t]
\centering
\caption{\textbf{Main result: four-condition ablation plus oracle}
($10$ catastrophic chains, $5$ seeds). \Lzero{}/\Lone{}/\Ltwo{} are success through one, two, and all three skills (the full chain); \Ltwo{} is our primary metric, seed-averaged per chain. $p$ is a sign-flip permutation test vs.\ Base over chains ($^{*}p{<}0.05$; $^{\dagger}p{<}0.10$), uncorrected across conditions, with the full-vs-base test pre-specified.}
\label{tab:main}
\scriptsize
\setlength{\tabcolsep}{3pt}
\begin{tabular}{lcccc}
\toprule
Method & \Lzero{} & \Lone{} & \Ltwo{} & $\Delta$\Ltwo{} vs.\ Base ($p$) \\
\midrule
Base (no recovery)            & $0.885$ & $0.343$ & $0.076$ & n/a \\
Base ($+$re-home/retry budget)& $0.882$ & $0.432$ & $0.129$ & $+0.053$ (step effect) \\
Recovery only                 & $0.876$ & $0.459$ & $0.175$ & $+0.099$, $p{=}0.094^{\dagger}$ \\
Prevention only               & $0.916$ & $0.404$ & $0.114$ & $+0.038$, $p{=}0.188$ \\
Prevention ($\sim$6k seam demos)& $0.898$ & $0.353$ & $0.117$ & $+0.041$, $p{=}0.062^{\dagger}$ \\
\textbf{Recovery $+$ Prev.}   & $0.906$ & $0.496$ & $\mathbf{0.265}$ & $\mathbf{+0.189}$, $\mathbf{p{=}0.047^{*}}$ \\
Oracle (privileged reset)     & $0.892$ & $0.673$ & $0.518$ & $+0.442$, $p{=}0.002^{*}$ \\
\bottomrule
\end{tabular}
\end{table}

Per scene, the full system reaches \Ltwo{} levels of $S1{:}\,0.29$, $S2{:}\,0.34$, $S4{:}\,0.16$; the S4 level sits at base ($0.19$), and one S4 chain (\texttt{ch3\_8}) regresses below base under the full system ($0.27{\to}0.17$; Tab.~\ref{tab:whenrestore}), a false-alarm cost we report rather than average away.

\textbf{Generalization to held-out seam orderings.} Seven of the ten chains
(\texttt{ch3\_11}--\texttt{20}) are held-out reorderings of the same skills, producing
novel seam transitions never used to design the system. 
The learned system improves \Ltwo{} $0.066\!\to\!0.244$ there ($3.7\times$, matching overall; too small alone for significance at $p{=}0.125$, but the oracle corroborates, $0.066\!\to\!0.499$). 
The subset was fixed by a base threshold (\Ltwo{}$\,\le 0.30$) before any recovery
On all $20$ chains base and oracle reach $0.286$ and $0.600$, so the phenomenon holds beyond
it, reflecting difficulty, not post-hoc selection.

\begin{table}[t]
\centering\footnotesize
\setlength{\tabcolsep}{4pt}
\caption{When to restore vs.\ retry: per-chain \Ltwo{}, \emph{recovery action isolated}
(matched re-home$+$retry vs.\ learned restoration, no prevention).}
\label{tab:whenrestore}
\begin{tabular}{llccc}
\toprule
Regime & Chain & base & retry & restore \\
\midrule
\multirow{3}{*}{Restore decisive}
 & \texttt{ch3\_2}  & $0.00$ & $0.00$ & $\mathbf{0.27}$ \\
 & \texttt{ch3\_3}  & $0.01$ & $0.12$ & $\mathbf{0.29}$ \\
 & \texttt{ch3\_11} & $0.00$ & $0.06$ & $\mathbf{0.09}$ \\
\midrule
\multirow{2}{*}{Retry suffices}
 & \texttt{ch3\_13} & $0.05$ & $\mathbf{0.31}$ & $0.25$ \\
 & \texttt{ch3\_16} & $0.20$ & $\mathbf{0.30}$ & $0.35$ \\
\midrule
\multirow{2}{*}{Restore hurts}
 & \texttt{ch3\_8}  & $0.32$ & $0.30$ & $0.21$ \\
 & \texttt{ch3\_12} & $0.25$ & $0.20$ & $0.10$ \\
\midrule
Unrecoverable & \texttt{ch3\_15,17,20} & $0.00$ & $0.00$ & $0.00$ \\
\bottomrule
\end{tabular}
\end{table}

\subsection{When restoration is the right fix}
\label{sec:whenrestore}
The diagnosis says \emph{what} to restore, but here we ask \emph{when} the deployed restoration actually helps, isolating the recovery action per chain (matched re-home$+$retry vs.\ learned restoration, no prevention; Tab.~\ref{tab:whenrestore}).
The outcome tracks one variable: our restorer acts only on the \emph{drawer}, so what matters is whether the seam's blocking displacement is a drawer that the downstream skill does not itself operate.
(i)~An off-canonical drawer blocks a non-drawer skill (a bowl onto the cabinet top while a drawer is stuck open, \texttt{ch3\_2,3}): restoration clears the obstruction and is decisive ($0\!\to\!0.28$), where a retry cannot move the drawer.
(ii)~The downstream skill \emph{is} a drawer operation (\texttt{ch3\_16,8,12}): restoring the drawer undoes the skill's own effect, so a retry is better and restoration can \emph{regress} an already-working chain ($0.32\!\to\!0.21$).
(iii)~The blocker is a displaced free \emph{object} (\texttt{ch3\_15,17,20}): the drawer restorer has no matching action and the chain is unrecovered.
The finding is that restoration helps exactly when a \emph{restorable} component blocks a skill that does not operate it. 
We tested the prescribed fix directly: a learned object restorer reaches $0.65$ at the skill level in isolation but only \Ltwo{}$\,{\approx}\,0.10$ at the chain level (vs.\ $0.265$ for drawer restoration), so object-blocked seams remain open.

\subsection{Detector quality and deployability}
\label{sec:deploy}

\begin{table}[t]
\centering\footnotesize\setlength{\tabcolsep}{5pt}
\caption{Detector accuracy and cost (pooled seam-window AUROC).}
\label{tab:detector}
\begin{tabular}{llcc}
\toprule
 & Encoder / model & AUROC & Latency \\
\midrule
\multirow{4}{*}{Ours} & DINOv3-B/16 (final) & \textbf{0.903} & $6.1$~ms \\
 & DINOv3-L/16 & $0.741$ & --- \\
 & DINOv3-S/16 & $0.739$ & --- \\
 & Wan-VAE (earlier) & $0.782$ & $38$~ms \\
\midrule
\multirow{2}{*}{Baselines$^{\dagger}$} & Robometer (4B PRM) & $0.746$ & --- \\
 & GVL (Qwen3-VL, 8B) & $0.718$ & --- \\
\bottomrule
\end{tabular}
\par\smallskip
\begin{minipage}{\columnwidth}\footnotesize
$^{\dagger}$Matched protocol vs.\ our earlier Wan-VAE head ($0.782$); the DINOv3-B head
($0.903$) is final ($\sim$1M params over a frozen encoder).
\end{minipage}
\end{table}

\textbf{Detector.} Detection must be cheap enough for a $20$~Hz loop and accurate
enough not to bottleneck recovery, so we compare encoders and external
progress/value models (Table~\ref{tab:detector}). A $\sim\!1$M-parameter progress
head over a frozen DINOv3-B/16 encoder gives a $6\times$ latency reduction over our
earlier Wan-VAE encoder and leads larger process-reward and VLM value baselines. In-loop, the better offline detector
reaches end-to-end parity with the video-model monitor, so the loop is limited by
recovery rather than detection. Substituting ground-truth seam timing for the learned
monitor reaches \Ltwo{}$\,0.226$, below the selective monitor's $0.265$ and far below
the $0.518$ oracle teleport, so better detection timing does not help, making the residual
gap one of restoration quality.

\begin{figure}[t]
\centering
\subfloat[Learned restoration skill\label{fig:rlpd}]{%
  \includegraphics[width=0.48\linewidth]{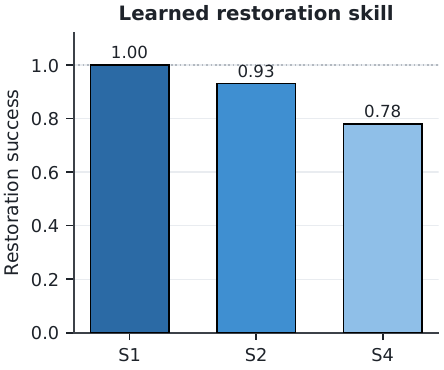}}
\hfil
\subfloat[Camera-only student\label{fig:student}]{%
  \includegraphics[width=0.48\linewidth]{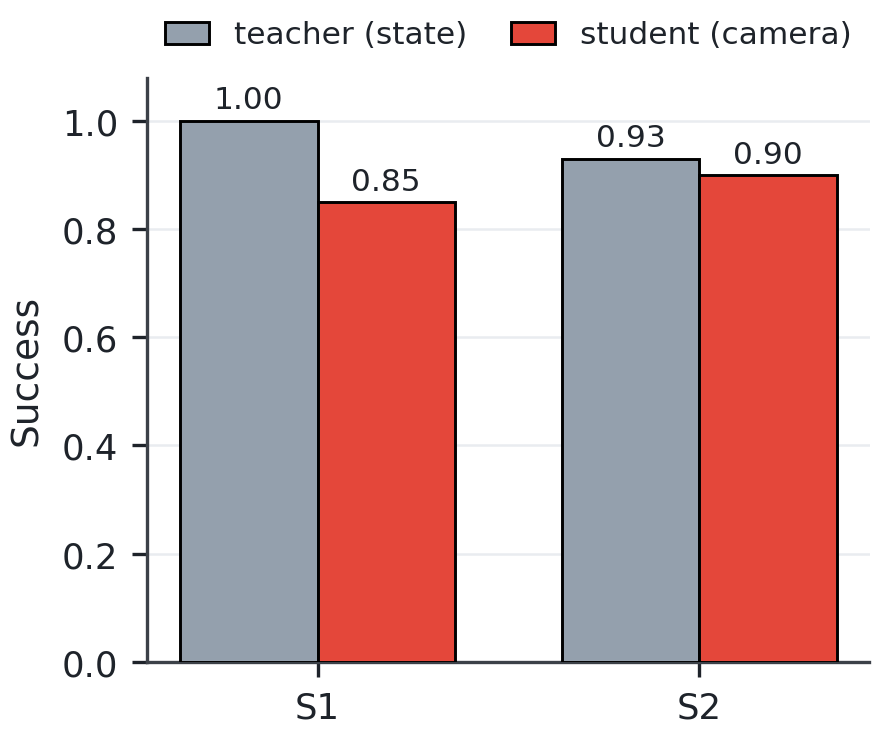}}
\caption{(a)~The learned restoration skill reaches $1.00/0.93/0.78$ (S1/S2/S4) at
the skill level. (b)~A distilled camera-only student matches the privileged
teacher, removing simulator state from the loop.}
\label{fig:skill}
\end{figure}

\textbf{Deployability.} The camera-only distilled student matches the privileged
teacher ($0.85/0.90$; Fig.~\ref{fig:student}), and the deployment controller runs
the entire detect, restore, resume loop on RGB and proprioception with zero
simulator state, validated in simulation through the same interface a real robot
would implement. The loop is real-time: the DINOv3 monitor costs $6.1$~ms/frame
(sampled every third step) and the restoration student $6$~ms/step on CPU (less on
GPU), both within the $50$~ms budget of the $20$~Hz control loop. Consuming only RGB and proprioception, the deployed loop is simulator- and embodiment-agnostic at test time. This transfers to any platform exposing these observations. The restoration skill is still \emph{trained} per scene in simulation, which the hardware study examines.

% ---- figure grid: rl curves + student/teacher --------------------------

\subsection{The monitor transfers to a real robot}
\label{sec:realrobot}
The simulation monitor runs unchanged on hardware, but its limit there is observability. We ran the loop on a Franka arm driven by a served, LoRA-fine-tuned $\pi_{0.5}$~\cite{pi05_2025} vision-language-action policy over three object-rearrangement chains (nine skills spanning shared-object, picking, and placing seams), $15$ trials per chain, with place-only
restoration under a matched step budget (both conditions get $200$ steps per skill, with
restoration counted inside it). The hardware restoration is a language-prompted
$\pi_{0.5}$ behavior, not the distilled camera-only student of Sec.~\ref{sec:restore}.
The progress monitor (Sec.~\ref{sec:detect}), was applied to the real exterior camera unchanged from simulation.
The live rule fires when max progress stays below $\tau{=}0.45$ by a mid-skill deadline.

\textbf{Exterior vision limits detection.} Scored in-sample the monitor looks near-perfect,
but that estimate is leaked (the head was fit on rollouts of the same objects it
scores). 
When scoring a new
episode with known objects and on unseen
objects, the deployment's AUROC falls to the range in Table~\ref{tab:realrobot}: wide at $n{=}45$ and
not separable from chance, so we report weak transfer and make no significant-detection
claim. 
The gap is observability, since with exterior vision alone a failed grasp or dropped
object is often invisible, even if the place motion still completes on camera. 
Gripper width and proprioception were not logged, making wrist and gripper sensing as the concrete fix (Sec.~\ref{sec:limitations}).

\textbf{Closing the loop is a proof of concept} (Fig.~\ref{fig:system}). At matched
detector flags, restoration recovers otherwise-terminal place attempts that the
baseline never recovers (Table~\ref{tab:realrobot}), which is an effect of the restore action
that does not depend on detector quality, since both conditions use the same live
flags. Per-skill success is otherwise within noise at $N{=}15$; we report the loop as
feasible and claim no success-rate improvement, since the bottleneck is observability.  

\begin{table}[t]
\centering\footnotesize\setlength{\tabcolsep}{4pt}
\caption{Real-robot study: Franka + served $\pi_{0.5}$, three chains, $15$ trials each.}
\label{tab:realrobot}
\begin{tabular}{lcc}
\toprule
\textbf{Detection} (exterior cam) & AUROC & $95\%$ CI \\
\midrule
Leave-one-episode-out & $0.625$ & $[0.46,0.79]$ \\
Leave-one-chain-out   & $0.672$ & $[0.51,0.83]$ \\
\midrule
\textbf{Recovery} (matched flags) & Recovered & \\
\midrule
Baseline (no restore) & $0/19$ & \\
$+$ restoration       & $\mathbf{3/21}$ & \\
\bottomrule
\end{tabular}
\par\smallskip
\begin{minipage}{\columnwidth}\footnotesize
Leave-one-episode-out $=$ known objects; leave-one-chain-out $=$ unseen objects.
Per chain: RC1 $1/7$, RC2 $1/8$, RC3 $1/6$. Detection catches $50\%$ of failures at a
$26\%$ false-alarm rate; at $n{=}45$ neither AUROC separates from chance
(Fisher $p{\approx}0.12$).
\end{minipage}
\end{table}

% ============================================================================
\section{Discussion}
\label{sec:discussion}
% ============================================================================
\textbf{One mechanism explains the results.} Seam failure is an off-support state
rather than a miscalibrated policy, so only restoring the state clears it. This is why
searching harder (best-of-$K$), swapping the policy (Diffusion Policy), or planning
(a world model) all leave the seam unmoved. The ablation says \emph{what} to restore:
articulation and objects are complementary, and any single-component fix reaches only
$0.13$--$0.27$ of the $0.52$ oracle (Table~\ref{tab:ablation}).

\textbf{The open problem is restoration quality.} Even where restoration applies it recovers only half the oracle gap (Table~\ref{tab:main}), and the object restorer we tried does not yet transfer (Sec.~\ref{sec:whenrestore}). 
We \emph{hypothesize} the limiting factor is re-entry timing (when to hand control back) rather than component coverage: the monitor can re-fire mid-restoration, and a fixed hand-back tolerance is a crude proxy for ``restored enough to resume.'' A learned re-entry value function, plus enabling restoration by skill type (Sec.~\ref{sec:whenrestore}), are the natural next steps.

% ==========================    ==================================================
\section{Conclusion and Future Work}
\label{sec:conclusion}
% ============================================================================
We diagnosed long-horizon seam collapse as Observation-Space Shift, localized it
causally to displaced articulation and secondary objects whose effects interact,
and showed, through controls that fail from the same states, that the test-time-compute,
stronger-base-policy, and world-model interventions we tested do not recover the seam,
whereas restoration does. Our fully learned, non-privileged detect–restore–resume system improves full-chain success $3.5\times$ over the frozen base
($p{=}0.047$), reaching $51\%$ of a privileged oracle from cameras and
proprioception alone.

\noindent \textbf{Limitations and future work.}
\label{sec:limitations}\label{sec:future}
Our quantitative core is in simulation. 
On hardware the exterior-camera monitor is limited by observability, since a failed grasp is invisible without wrist and gripper sensing, the clearest next step. 
Restoration is defined only for drawer seams, so chains whose second seam displaces a free object stay unrecovered and need an object-restoration skill. 
Results use a single base-policy seed and an uncorrected $n{=}10$ test, and the absolute \Ltwo{} ($0.265$) is modest. 
Scaling seam data $10\times$ within the same distribution does not help ($0.117$). Whether broader training distribution randomization would is untested.

\section*{Acknowledgment}
The authors used Anthropic's Claude, an AI assistant, in preparing this manuscript. 
It was used for: 
(i) language editing and clarity/concision revision of author-written text throughout; 
(ii) Refining passages and figure captions in the Introduction, Method, and Experiments under author direction. 
All AI-assisted text reviewed, verified, and edited by the authors, who take full responsibility for the content and claims.

% ============================================================================
% ---- bibliography ----------------------------------------------------------
% ============================================================================
\bibliographystyle{IEEEtran}
\bibliography{references}

\end{document}